\documentclass[11pt]{article}

\usepackage[margin=1in]{geometry}
\usepackage{times,natbib}

\usepackage[utf8]{inputenc} 
\usepackage[T1]{fontenc}    
\usepackage[hidelinks]{hyperref} 
\usepackage{url}            
\usepackage{booktabs}       
\usepackage{amsfonts}       
\usepackage{nicefrac}       
\usepackage{microtype}      
\usepackage{xcolor}         

\usepackage{graphicx}
\usepackage{xspace}
\usepackage{caption}
\usepackage{enumitem}
\usepackage{amsmath}
\usepackage{wrapfig}
\usepackage{subcaption}
\usepackage{placeins}
\usepackage{float}
\usepackage{tikz}
\usetikzlibrary{arrows.meta,positioning}

\newcommand{\ours}{CAP\xspace}

\title{CAP: Learning to Optimize from Sparse Observations of Related Tasks}

\author{Lin Fu \quad Siwei Fu \quad Dong Ni\\[0.35em]
Zhejiang University\\[0.35em]
\texttt{22447022@zju.edu.cn}}
\date{}

\begin{document}
\raggedbottom

\maketitle

\begin{abstract}
In experiment-driven black-box optimization, each evaluation can consume materials, equipment time, and other resources, limiting the number of experiments available for a new task. Records from completed related optimization campaigns could help, but learning a sequential query policy remains challenging when task-specific experience consists of only a few sparse histories. We propose CAP (Candidate Acquisition Policy), which models shared structure and task-specific variation in fixed historical records to sample interactive practice functions. On these functions, CAP learns a feedback-conditioned query policy aimed at final optimization performance under a fixed evaluation budget. At deployment, its parameters remain fixed while each new observation informs the next query. Across five synthetic task families spanning 6 and 20 dimensions and two engineering simulation benchmarks, BOPTEST and SU2, CAP attains the lowest median final-budget value among the reported methods on four synthetic families and both engineering benchmarks. These results indicate that sparse records from completed related tasks can support adaptive query decisions on new tasks, making more effective use of limited, costly experimental evaluations.

\end{abstract}

\section{Introduction}
Experiment-driven black-box optimization seeks high-quality solutions with a limited number of objective evaluations. Each evaluation may consume materials or equipment time, or require a time-consuming engineering simulation, making effective use of every query on a new task important~\citep{shields2021bayesian,frazier2018tutorial}. Bayesian optimization (BO) uses observations and predictive uncertainty to guide subsequent queries under such budgets~\citep{frazier2018tutorial}. Yet a new task need not start entirely from scratch: completed optimization campaigns on related tasks leave records that may contain useful experience. This paper asks how that experience can be turned into effective query decisions on a new task.

We study a setting with limited historical experience: only sparse optimization records from a small number of related tasks are available, and these completed tasks cannot be evaluated further. The goal is to learn a query policy from the records that uses incoming feedback to optimize a new, related task within a fixed evaluation budget. Related tasks can share useful structure without having identical response functions, so configurations that worked well in the past cannot simply be reused. A policy must draw on historical experience while continually revising its decisions in light of observations from the current task.

Transfer BO methods can use historical observations to improve predictions or query selection on a new task~\citep{wistuba2021few}. Learned optimizers go further by learning query behavior from related tasks~\citep{volpp2020metalearningacquisitionfunctionstransfer,hsieh2021reinforced}. Training such behavior requires tasks or approximations that can answer newly proposed queries; these interactions need not be physical experiments. In our setting, however, task-family-specific experience is limited to sparse, fixed records of completed campaigns. The records cannot directly answer queries proposed during policy training, and the available observations do not uniquely determine the response surfaces of related tasks. The central challenge is therefore not merely to identify promising locations from historical data, but to construct grounded interactive practice within this information boundary and learn a sequential strategy that adapts to feedback on a new task.

We propose \ours{} (Candidate Acquisition Policy). \ours{} models structure shared across historical tasks alongside task-specific variation, then samples interactive practice functions from the fitted distribution. These functions provide sequential optimization experience without further evaluations of the historical tasks. On them, \ours{} learns candidate-selection behavior aimed at final performance under a fixed evaluation budget. When deployed on a new related task, the policy parameters remain fixed: decisions use the growing observation context and candidate information from a frozen pretrained predictor, and each new evaluation informs the next query. Historical experience is thus distilled into reusable query behavior, while adaptation to a particular target task occurs through feedback rather than policy fine-tuning.

We evaluate \ours{} on five synthetic task families spanning 6 and 20 dimensions and on two engineering simulation benchmarks, building control with BOPTEST~\citep{blum2021building} and airfoil design with SU2~\citep{economon2016su2}, against single-task, transfer, and learned optimizers. \ours{} attains the lowest median final-budget value among the reported methods on four synthetic families and both engineering benchmarks; FSAF leads on Levy20. These results indicate that sparse records from completed related tasks can support adaptive queries on new tasks, making more effective use of limited, costly evaluations.

Our main contributions are:
\begin{itemize}[leftmargin=1.5em]
    \item We develop a way to construct interactive practice functions from sparse, fixed records of related tasks, enabling sequential query-policy training from completed campaigns.
    \item We introduce \ours{}, which combines record-derived practice with policy learning aimed at final performance under a fixed budget. At deployment, the policy adapts its decisions through the growing observation context without updating its parameters.
    \item We evaluate the complete optimizer across synthetic task families with different landscapes and dimensions and two engineering simulation benchmarks, demonstrating its final-budget optimization performance.
\end{itemize}

\section{Related Work}

Work on query efficiency under limited evaluation budgets either improves search and acquisition rules or uses experience from related tasks to learn predictions and decisions. In the former direction, TuRBO organizes search within local trust regions~\citep{eriksson2019scalable}, while FunBO discovers reusable acquisition functions~\citep{aglietti2024funbodiscoveringacquisitionfunctions}. \ours{} focuses on a more constrained setting in the latter direction: only fixed records from a small number of related tasks are available, yet the goal is to train a policy that selects queries using feedback from a new task. We discuss related work on cross-task transfer, query policy learning, and in-context prediction.

\subsection{Transfer Learning for Bayesian Optimization}
Transfer BO typically uses historical experience in the surrogate or acquisition decisions for a new task. RGPE weights GP models according to the ranking performance of historical models on current-task observations~\citep{feurer2018scalable}; ABLR relates regression models across tasks through a shared representation~\citep{perrone2018scalable}; and FSBO meta-learns a deep-kernel surrogate that adapts to new-task observations~\citep{wistuba2021few}. At the acquisition level, TAF combines historical-task information with expected improvement on the new task~\citep{wistuba2018scalable}. \ours{} also uses cross-task relatedness, but the historical task model generates practice functions for policy training rather than predicting candidate responses at deployment; training incorporates historical experience into the policy parameters.

\subsection{Learning Query Policies for Black-Box Optimization}
MetaBO trains a neural acquisition function with reinforcement learning~\citep{volpp2020metalearningacquisitionfunctionstransfer}; FSAF connects acquisition functions to action values~\citep{hsieh2021reinforced}; and NAP jointly learns predictive representations and acquisition decisions~\citep{maraval2023end}. OptFormer takes another route, learning query behavior from existing hyperparameter optimization trajectories~\citep{chen2022learninguniversalhyperparameteroptimizers}.

Interactive training of the first three methods requires responses at points selected by the policy: MetaBO uses evaluable tasks or inexpensive approximations, FSAF uses sampleable functions, and NAP uses candidate responses available from source tasks. In our setting, only sparse records from a small number of completed tasks are available; if a policy selects an unrecorded point, the records cannot provide feedback, so subsequent training trajectories cannot proceed as in the original setups. Thus, the limitation is not in these policy models themselves, but in the inability to meet their original training conditions directly from fixed records. \ours{} constructs queryable computational practice from the records, then learns a feedback-conditioned policy aimed at final performance under a fixed budget.

\subsection{Learned Priors and In-Context Prediction}
Prior-data fitted networks (PFNs) make predictions conditioned on an observation context~\citep{muller2021transformers}; TabPFN applies this approach to prediction on small tabular datasets~\citep{hollmann2025tabpfn}, and PFNs4BO studies its use in Bayesian optimization~\citep{muller2023pfns4bo}. \ours{} uses a frozen pretrained predictor to provide candidate features for the current task and trains the query policy separately. This predictor differs from the practice-function model fitted to historical records: the former supports current decisions, whereas the latter provides an interactive environment only during training.


\section{Method}
\label{sec:method}

\subsection{Problem Setting and Overview}
\label{sec:problem}
\label{sec:overview}

We consider related black-box minimization tasks with comparable decision variables, mapped to a common domain $\mathcal X=[0,1]^d$. The available historical experience consists of fixed optimization records from $K$ completed tasks,
\begin{equation}
\mathcal D=\{D_h\}_{h=1}^{K},\qquad
D_h=\{(x_{h,i},y_{h,i})\}_{i=1}^{n_h}.
\end{equation}
The tasks need not have identical response functions. Training may use these records and quantities derived from them, but cannot obtain further evaluations of the historical objectives or interact with a new target task.

On a new related task $f_\star$, an initial design of $n_0$ evaluations is followed by $H=B-n_0$ adaptive queries within a total budget $B$. Let $O_t$ contain the initial observations and the first $t$ adaptive evaluations, and let $b_t=\min_{(x,y)\in O_t}y$ be the best value observed so far. The objective is a small terminal value $b_H$; when a task optimum is available to the evaluator, it can also compute terminal simple regret. Neither the optimum nor unqueried target responses are available to the policy. Historical record length $n_h$ and target budget $B$ have distinct roles, even when an experimental protocol assigns them the same numerical scale.

\begin{figure}[!ht]
\centering
\includegraphics[width=\textwidth]{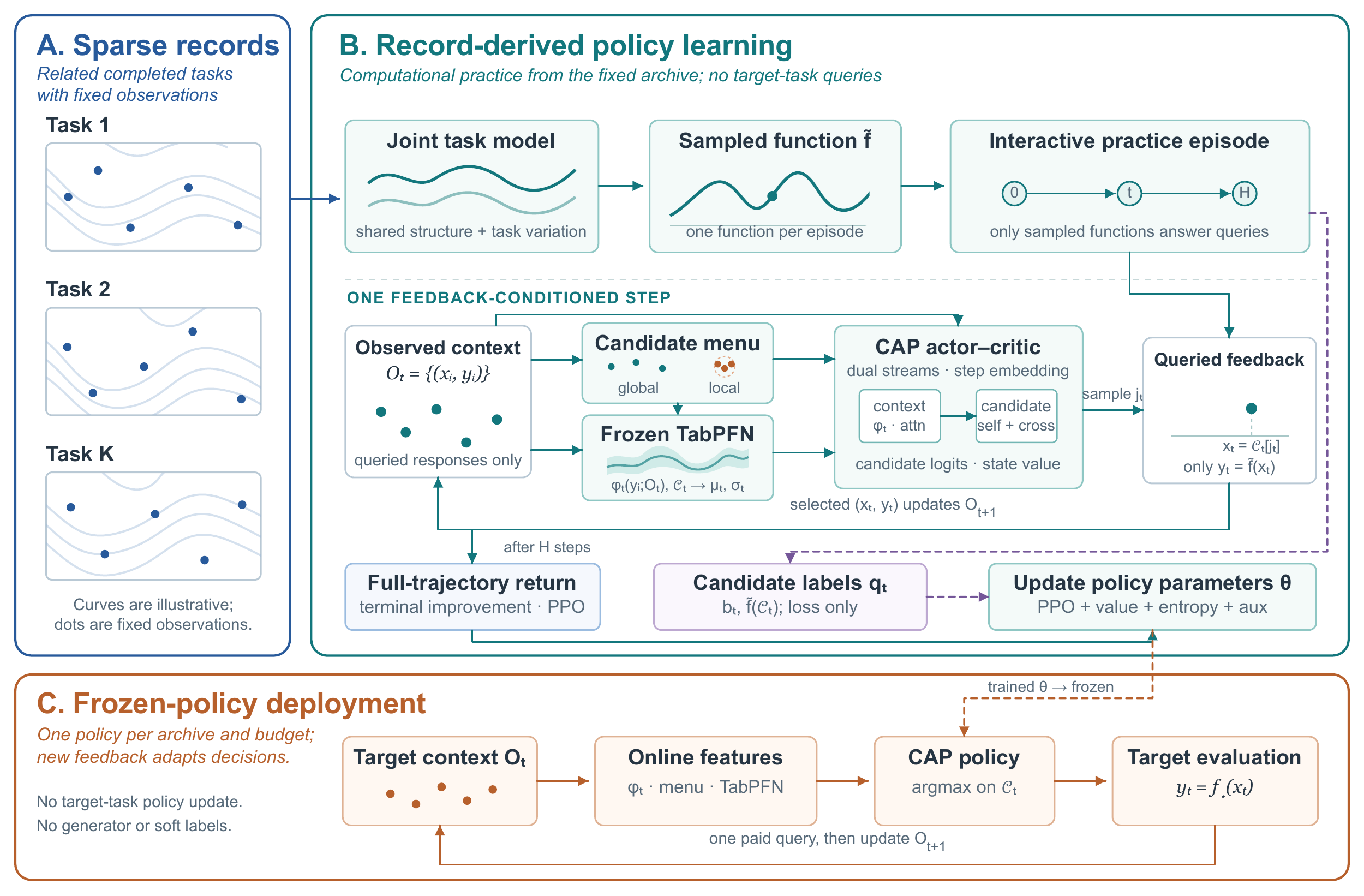}
\caption{Overview of \ours. Fixed historical records fit an approximate distribution of coherent practice functions on which complete trajectories train a candidate-selection policy. At each step, the menu and represented observations inform a frozen, externally pretrained TabPFN and the actor--critic; only the selected response updates the observation context. Full trajectories yield PPO returns, while responses of all practice candidates supply auxiliary labels $q_t$ only to the training loss. At deployment, one frozen policy per historical archive and budget condition uses current-task feedback and pays for only the selected target query.}
\label{fig:overview}
\end{figure}

\ours{} turns the fixed records into interactive computational practice and learns sequential query behavior over complete budgets. For a given historical archive and budget condition, training yields one policy whose parameters remain frozen when deployed on new tasks associated with that archive. The method is fitted and trained separately for different archives; a single set of weights is not assumed to transfer across unrelated task families. Adaptation to a new task comes from its growing observation context and the resulting query decisions, not from retraining the policy. Figure~\ref{fig:overview} summarizes this process.

\subsection{History-Conditioned Practice Functions}
\label{sec:synthetic_episodes}

Fixed records provide responses only at previously queried locations and cannot directly answer new queries proposed during policy training. To obtain complete sequential training trajectories, we fit an approximate, history-conditioned task distribution and sample
\begin{equation}
\widetilde f\sim\widehat{\mathcal P}_{\mathcal D}.
\label{eq:practice_distribution}
\end{equation}
Each draw defines a function path fixed throughout an optimization episode. Interacting with it is computation derived from the records, not another evaluation of a historical or target objective.

\paragraph{Shared structure and task variation.}
For the continuous part of the responses, we model historical observations with shared spatial structure and task-specific variation:
\begin{equation}
\begin{aligned}
z_{h,i}:=T_\eta(y_{h,i})
&=\beta+g(x_{h,i})+a_h+u_h(x_{h,i})+\epsilon_{h,i},\\
g&\sim\mathcal{GP}(0,k_s),\qquad
u_h\overset{\mathrm{iid}}{\sim}\mathcal{GP}(0,k_u),\\
a_h&\overset{\mathrm{iid}}{\sim}\mathcal N(0,v_a),\qquad
\epsilon_{h,i}\sim\mathcal N(0,\sigma_\epsilon^2).
\end{aligned}
\label{eq:joint_task_model}
\end{equation}
Here $T_\eta$ transforms responses for historical modeling, $\beta$ is a common mean, $g$ captures structure shared across tasks, $u_h$ captures task-specific spatial variation, and $a_h$ is an optional task-level offset. The kernels $k_s$ and $k_u$ describe the shared and task-specific processes, respectively; $\epsilon_{h,i}$ is a fitting-noise term for historical observations. All $K$ records contribute to the joint fit. In a new practice task, the shared structure is conditioned on the records, while its task-specific process and offset are drawn anew. A practice function therefore need not reproduce any particular historical response surface.

We approximate prior function paths with random Fourier features and condition whole paths on the historical observations~\citep{rahimi2007random,wilson2020efficientlysamplingfunctionsgaussian}. The path and any applicable discrete response component are fixed at the start of an episode; repeated queries at the same location do not resample fitting noise. The resulting $\widehat{\mathcal P}_{\mathcal D}$ is an approximation constrained by sparse records, not an exact recovery of the true task distribution. Response-transformation families and whether a discrete component is enabled are specified for each experimental condition before target-task evaluation; their numerical parameters are fitted only from the corresponding historical records. These response-modeling choices do not change the subsequent policy architecture or training objective. Path construction and the response choices are detailed in Appendix~\ref{app:method_details}.

\subsection{Feedback-Conditioned Candidate Acquisition}
\label{sec:cap_policy}

The historical archive influences query behavior through the trained policy parameters, but is not supplied as a direct input at each target-task decision. The policy instead uses the current task's evaluated observations, candidate locations, predictions based only on those observations, and the decision stage. The practice generator's latent variables and unqueried responses are unavailable.

\paragraph{Current-task response representation.}
To separate response modeling for practice-task generation from online decision features, define a context-dependent representation
\begin{equation}
\bar y_{t,i}=\phi_t(y_i;O_t),\qquad (x_i,y_i)\in O_t.
\label{eq:online_response_representation}
\end{equation}
Unlike $T_\eta$, which is used to construct practice functions, $\phi_t$ depends only on the current task's evaluated observations and is recomputed after every query. The synthetic experiments use the identity representation; the engineering experiments use within-context normal scores based on response ranks, so wide response ranges do not compress input differences among favorable observations. Each representation is held fixed for its training condition and used identically at deployment; it is not selected using target-task outcomes. The representation affects only predictor and policy inputs. Incumbents, training returns, auxiliary labels, and reported objective values retain the original response units.

\paragraph{Candidate menu and online predictor.}
At step $t$, a fixed-size menu $\mathcal C_t=\{x_{t,j}\}_{j=1}^{M}$ combines a persistent global Sobol pool with local candidates regenerated around spatially separated, favorably observed locations. Selected global points are removed from the pool. Local sampling uses a predefined budget-dependent radius schedule, while its centers respond to the observations. The policy learns which available candidate to query; it does not learn the menu's allocation or radius rule.

A frozen, externally pretrained TabPFN-v2 regressor predicts candidate means $\mu_t(x)$ and standard deviations $\sigma_t(x)$ from $\{(x_i,\bar y_{t,i})\}_{(x_i,y_i)\in O_t}$~\citep{hollmann2025tabpfn}. These predictions are in the $\bar y$ representation space. TabPFN supplies current-task decision features, whereas the historical model in Eq.~\eqref{eq:joint_task_model} supplies training environments. Task-family-specific practice and policy learning come from the fixed records, distinct from TabPFN's external pretraining.

\paragraph{Query decisions and feedback.}
A dual-stream attention actor--critic encodes observation and candidate tokens separately and uses candidate-to-context cross-attention to form feedback-conditioned candidate representations. A learned step embedding indicates the stage of the fixed query horizon. The actor assigns a logit $\ell_{\theta,t,j}$ to each candidate and defines
\begin{equation}
\pi_\theta(j\mid s_t)=
\frac{\exp(\ell_{\theta,t,j})}{\sum_{k=1}^{M}\exp(\ell_{\theta,t,k})},
\label{eq:cap_policy}
\end{equation}
where $s_t$ comprises the represented observations, candidate features, and the stage within the fixed budget. The critic shares the encoder and estimates state value. Training samples an action from $\pi_\theta$; deployment selects its highest-scoring candidate. Only that candidate's response is added to $O_{t+1}$, after which the representation, predictions, and menu are rebuilt. The policy can thus revise subsequent decisions in response to new feedback, but learns selection within the predefined menu rather than its allocation or radius rule.

\subsection{Learning from Terminal Returns and Candidate Responses}
\label{sec:ppo_training}

Practice functions provide two complementary training signals: terminal improvement over complete query trajectories and additional responses at the current candidates. The former targets fixed-budget optimization; the latter provides denser supervision at visited states, but reflects immediate response quality only.

\paragraph{Terminal objective.}
For each practice function $\widetilde f$, a fixed reference set determines a positive response scale $C_{\widetilde f}$ before the episode. The scale remains constant within the episode and is used only for training. With $b_t$ computed from raw practice responses, define
\begin{equation}
r_t=\frac{b_t-b_{t+1}}{C_{\widetilde f}},\qquad
\sum_{t=0}^{H-1}r_t=\frac{b_0-b_H}{C_{\widetilde f}}.
\label{eq:terminal_improvement}
\end{equation}
For a fixed practice function and initial design, the undiscounted return equals its scale-normalized terminal improvement; across functions, training optimizes the expectation of this quantity. No exact function optimum is needed for the reward.

We update the actor--critic with clipped PPO~\citep{schulman2017proximalpolicyoptimizationalgorithms}, using returns and advantages from complete sampled trajectories. A query without immediate improvement can therefore receive a learning signal through later improvements in its trajectory. Optimization and numerical settings are given in Appendix~\ref{app:method_details}.

\paragraph{State-adaptive candidate supervision.}
At each visited state, the sampled practice function can also answer queries for the other candidates. Their signed, immediate utilities are
\begin{equation}
u_{t,j}=\frac{b_t-\widetilde f(x_{t,j})}{C_{\widetilde f}}.
\label{eq:candidate_utility}
\end{equation}
We retain negative values so that candidates can still be ordered when none improves the incumbent. Sort utilities as $u_{t,(1)}\ge\cdots\ge u_{t,(M)}$ and form a soft label with a state-dependent temperature:
\begin{equation}
\begin{aligned}
\tau_t&=\max\{\alpha(u_{t,(1)}-u_{t,(k)}),\epsilon_\tau\},\\
q_{t,j}&=(1-\zeta)\frac{\exp(u_{t,j}/\tau_t)}
{\sum_{\ell=1}^{M}\exp(u_{t,\ell}/\tau_t)}+\frac{\zeta}{M}.
\end{aligned}
\label{eq:adaptive_labels}
\end{equation}
Here $k$ is the candidate rank used to set the temperature, $\alpha>0$ controls its scale, $\epsilon_\tau>0$ supplies a numerical floor, and $\zeta\in[0,1]$ mixes the labels with a uniform distribution. We detach $q_t$ from gradient computation and minimize
\begin{equation}
\mathcal L_{\mathrm{aux}}=-\mathbb E_{s_t}\sum_{j=1}^{M}
q_{t,j}\log\pi_\theta(j\mid s_t).
\label{eq:candidate_auxiliary}
\end{equation}
These labels describe immediate response quality, not the long-term information value of a query. They complement, rather than replace, the complete-trajectory objective.

\paragraph{Joint training and deployment.}
The training loss combines the clipped PPO policy loss, a value loss, entropy regularization, and candidate supervision:
\begin{equation}
\mathcal L=\mathcal L_{\mathrm{clip}}+c_V\mathcal L_V
-c_{\mathrm{ent}}\mathbb E_{s_t}\mathcal H(\pi_\theta(\cdot\mid s_t))
+\lambda_{\mathrm{aux}}\mathcal L_{\mathrm{aux}}.
\label{eq:joint_policy_loss}
\end{equation}
Here $\mathcal L_{\mathrm{clip}}$ is the PPO actor loss, $\mathcal L_V$ trains the critic to predict complete-trajectory returns, $\mathcal H$ denotes policy entropy, and $c_V,c_{\mathrm{ent}},\lambda_{\mathrm{aux}}$ are loss weights. Responses of unselected candidates are used only to construct training labels; they do not enter the policy state, TabPFN context, or subsequent observation trajectory. These extra evaluations are computational queries to practice functions, not target-task evaluations.

After training, deployment retains the frozen predictor, candidate construction, and one fixed policy for the historical archive and budget condition. It does not use the practice generator or candidate labels. Each adaptive step pays for only the selected target query and returns the best point observed within the budget.

\section{Experiments}
\label{sec:experiments}

We ask whether a policy learned from a small number of fixed historical optimization records can improve budget-limited optimization on new, related tasks. We first compare the complete optimizer across five synthetic task families with different landscapes and dimensions, then examine whether its performance extends to two engineering simulation benchmarks.

\subsection{Experimental Setup}
\label{sec:exp_setup}

\paragraph{Tasks and budgets.}
We evaluate five synthetic families---Hartmann6, Ackley6, Levy6, Ackley20, and Levy20---each with 20 historical tasks ($5d$ fixed observations per task) and 100 distinct target tasks; their construction is outlined in Appendix~\ref{app:synthetic_construction}. BOPTEST building control and SU2 airfoil design each have 20 historical and 20 target tasks~\citep{blum2021building,economon2016su2}. All methods use the same target tasks, shared initial design ($n_0=d/2$), and total budget ($B=5d$ evaluations) within each benchmark. \ours{} trains one policy per historical archive and freezes it on target tasks. Further protocol is in Appendix~\ref{app:response_configurations}; synthetic scoring is in Appendix~\ref{app:synthetic_results}.

\paragraph{Baselines.}
Single-task optimizers are EI, PI, UCB, RolloutEI-2, FunBO, and TuRBO; FunBO uses a discovered acquisition function and TuRBO uses local trust regions~\citep{aglietti2024funbodiscoveringacquisitionfunctions,eriksson2019scalable}. Record-based transfer methods are TAF-ranking, RGPE, ABLR, and FSBO~\citep{wistuba2018scalable,feurer2018scalable,perrone2018scalable,wistuba2021few}. FSAF, MetaBO, and NAP learn query policies through interaction~\citep{hsieh2021reinforced,volpp2020metalearningacquisitionfunctionstransfer,maraval2023end}. We also include publicly pretrained PFNs4BO as an external-prior reference~\citep{muller2023pfns4bo}. Each retains its native surrogate, candidate construction, and decision rule; objective-evaluation budgets are matched.

FSAF, MetaBO, and NAP need training tasks that answer new queries, but the historical records cannot. We therefore train them on the same record-derived, queryable practice-function distribution as \ours, within the same historical-information boundary. They share practice feedback, not \ours{}'s predictor, architecture, candidates, or loss. The distinct roles of \ours{}'s frozen TabPFN predictor and the externally pretrained PFNs4BO model are explained in Section~\ref{sec:cap_policy}.

\paragraph{Evaluation metric.}
For synthetic tasks, we report median normalized simple regret across targets, at the final budget and along the trajectory. Since engineering optima are unknown, we report median best-observed objective values in original units. Lower is better. Appendix~\ref{app:synthetic_results} details scoring and aggregation.

\subsection{Synthetic Task Families}
\label{sec:synthetic_results}

\begin{figure}[!ht]
\centering
\includegraphics[width=0.92\linewidth]{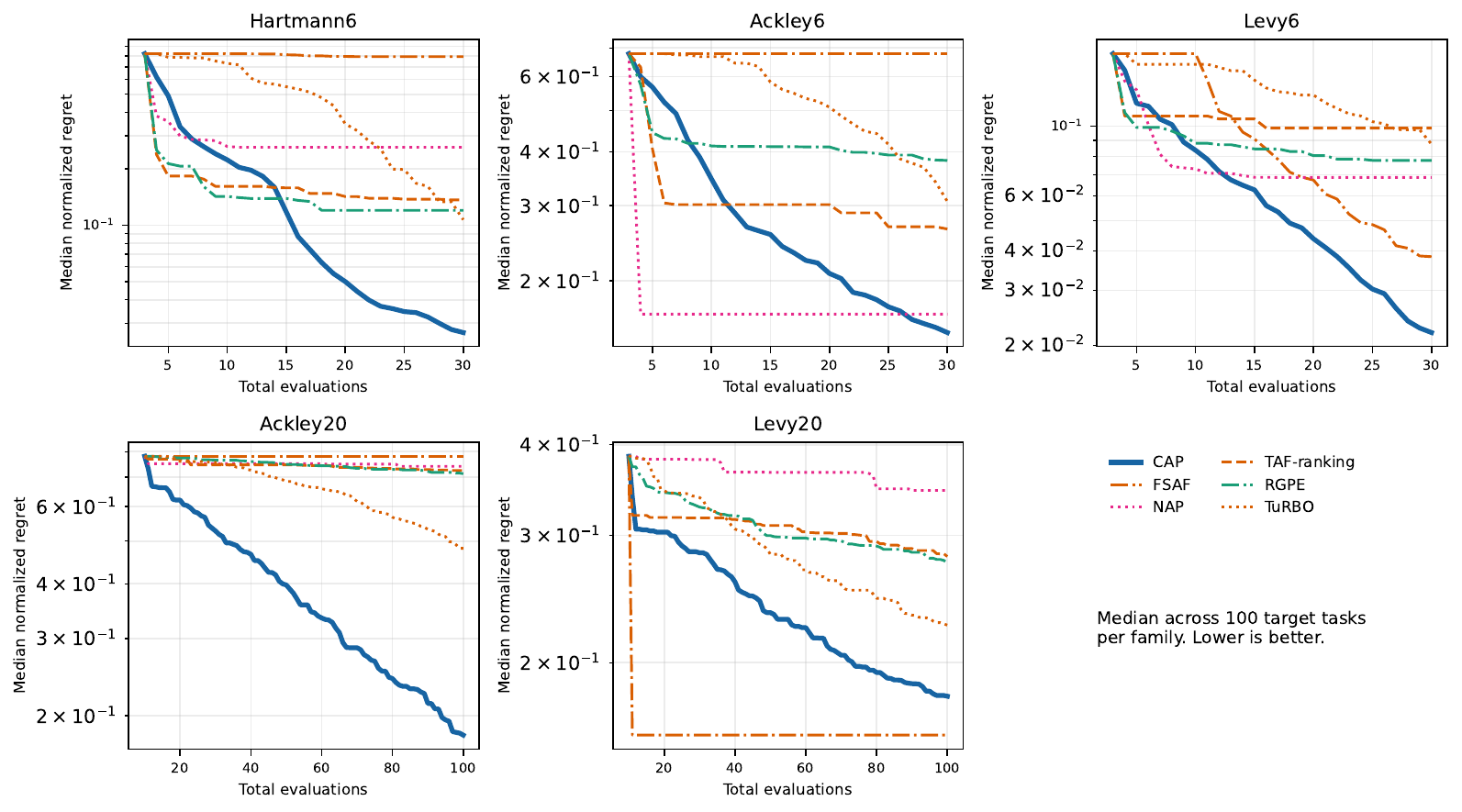}
\caption{Median normalized simple regret across 100 target tasks per family (lower is better). Curves begin after the initial design, included in the evaluation count; panel scales differ. Appendix~\ref{app:synthetic_results} shows all methods.}
\label{fig:synthetic_median_overview}
\end{figure}

Table~\ref{tab:synthetic_median_regret} compares final regret across all methods; Figure~\ref{fig:synthetic_median_overview} shows trajectories for a fixed cross-category selection. All trajectories appear in Appendix~\ref{app:synthetic_results}.

\begin{table}[!htbp]
\centering
\footnotesize
\caption{Final median normalized simple regret on 100 target tasks per family (lower is better; best in bold). PFNs4BO supports at most 18 dimensions and is omitted in 20D.}
\label{tab:synthetic_median_regret}
\begin{tabular*}{\linewidth}{@{\extracolsep{\fill}}lrrrrr@{}}
\toprule
Method & Hartmann6 & Ackley6 & Levy6 & Ackley20 & Levy20 \\
\midrule
\ours{} & \textbf{0.0267} & \textbf{0.1514} & \textbf{0.0220} & \textbf{0.1806} & 0.1798 \\
FSAF & 0.7936 & 0.6786 & 0.0384 & 0.7808 & \textbf{0.1589} \\
MetaBO & 0.1511 & 0.6052 & 0.0721 & 0.7808 & 0.3311 \\
NAP & 0.2606 & 0.1672 & 0.0686 & 0.7407 & 0.3460 \\
\midrule
TAF-ranking & 0.1365 & 0.2643 & 0.0985 & 0.7248 & 0.2804 \\
RGPE & 0.1201 & 0.3821 & 0.0778 & 0.7125 & 0.2747 \\
ABLR & 0.1765 & 0.5680 & 0.0616 & 0.7361 & 0.3212 \\
FSBO & 0.0900 & 0.2479 & 0.0639 & 0.6822 & 0.2308 \\
PFNs4BO & 0.1610 & 0.3717 & 0.0877 & N/A & N/A \\
\midrule
TuRBO & 0.1070 & 0.3065 & 0.0876 & 0.4816 & 0.2254 \\
EI & 0.4870 & 0.5060 & 0.1290 & 0.7167 & 0.2839 \\
PI & 0.4588 & 0.5230 & 0.1165 & 0.7155 & 0.2833 \\
UCB & 0.4840 & 0.5291 & 0.1372 & 0.7133 & 0.2872 \\
FunBO & 0.5010 & 0.5145 & 0.1219 & 0.7144 & 0.2840 \\
RolloutEI-2 & 0.5063 & 0.5350 & 0.1146 & 0.7225 & 0.2824 \\
\bottomrule
\end{tabular*}
\end{table}

\FloatBarrier

\ours{} has the lowest median final regret among the reported methods in four of the five families, including Ackley20 (Table~\ref{tab:synthetic_median_regret}). On Levy20, FSAF has the lowest median (0.1589), followed by \ours{} (0.1798). Thus the endpoint benefit spans different landscapes and dimensions, but does not extend to every baseline on every family. The trajectories in Figure~\ref{fig:synthetic_median_overview} show how the methods improve with target-task feedback, complementing the fixed-budget comparison in the table.

\subsection{Engineering Simulation Tasks}
\label{sec:physical_results}

Synthetic families allow controlled variation between related tasks, whereas the engineering experiments use specific simulator objectives and operating conditions. BOPTEST varies operating time windows within a shared building and control space; SU2 varies flight conditions within a shared airfoil-design space. BOPTEST has $d=6$ and a budget of 30 simulator evaluations per target task; SU2 has $d=10$ and a budget of 50. Policy training uses only the fixed historical records and their derived practice tasks, not simulator calls on target tasks.

\begin{figure}[H]
\centering
\includegraphics[width=0.90\linewidth]{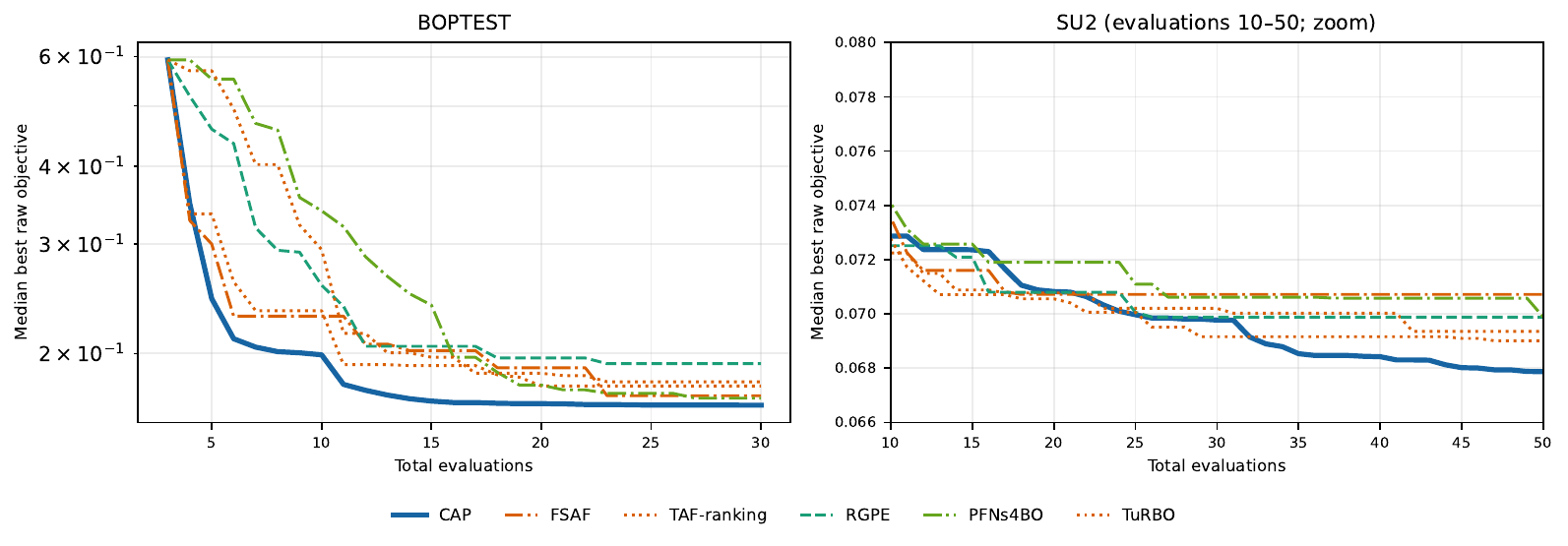}
\caption{Median best-observed objective versus simulator evaluations (lower is better). Vertical axes use each benchmark's original scale. The SU2 panel zooms to evaluations 10--50 and objective values 0.066--0.080; earlier evaluations still count toward the budget. Appendix~\ref{app:engineering_curves} shows all methods.}
\label{fig:physical_trajectories}
\end{figure}

Table~\ref{tab:physical_final} reports the median final best-observed objective for all reported methods on both benchmarks. Figure~\ref{fig:physical_trajectories} shows their progress within the simulation budget; trajectories for every reported method appear in Appendix~\ref{app:engineering_curves}.

\begin{table}[!b]
\centering
\scriptsize
\renewcommand{\arraystretch}{0.82}
\caption{Final median best-observed objective across 20 target tasks per engineering benchmark (lower is better). Values use each benchmark's original scale; bold marks the best result.}
\label{tab:physical_final}
\begin{tabular*}{0.88\linewidth}{@{\extracolsep{\fill}}lrr@{}}
\toprule
Method & BOPTEST (6D) & SU2 (10D) \\
\midrule
\ours{} & \textbf{0.1651} & \textbf{0.06787} \\
FSAF & 0.1710 & 0.07072 \\
MetaBO & 0.2078 & 0.07048 \\
NAP & 0.1878 & 0.07065 \\
\midrule
TAF-ranking & 0.1773 & 0.06935 \\
RGPE & 0.1926 & 0.06988 \\
ABLR & 0.4497 & 0.07018 \\
FSBO & 0.2786 & 0.07024 \\
PFNs4BO & 0.1695 & 0.06988 \\
\midrule
TuRBO & 0.1801 & 0.06901 \\
EI & 0.2057 & 0.06944 \\
PI & 0.1943 & 0.06963 \\
UCB & 0.1808 & 0.07007 \\
FunBO & 0.1935 & 0.06952 \\
RolloutEI-2 & 0.1814 & 0.06950 \\
\bottomrule
\end{tabular*}
\end{table}

\ours{} obtains the lowest median final best-observed objective among the reported methods on both BOPTEST and SU2 (Table~\ref{tab:physical_final}). The trajectories show how this performance develops across simulator evaluations (Figure~\ref{fig:physical_trajectories}). Together with the synthetic results, these experiments demonstrate its budget-limited optimization performance across the evaluated task families.

The two engineering benchmarks both show modest final-median margins. On BOPTEST, the final medians of \ours{} (0.1651), PFNs4BO (0.1695), and FSAF (0.1710) are close. Several methods improve rapidly early in the budget and approach similar values near its end. On SU2, \ours{} reaches 0.06787 after 50 evaluations, while the next lowest median among the reported methods is TuRBO's 0.06901. The zoomed trajectory shows a continuing but small late-budget improvement. These comparisons are within each benchmark: BOPTEST and SU2 retain their own objective scales, and the reported medians do not imply a win on every target task.

\section{Conclusion}
We introduced \ours{} to turn fixed, sparse records of completed related tasks into interactive computational practice and learn a sequential query policy for a fixed evaluation budget. On a new task, the policy requires no further training: it adjusts subsequent queries as observations accumulate. Across five synthetic families and two engineering simulation benchmarks, \ours{} achieves the lowest median final-budget value among the compared methods on four synthetic families and both engineering benchmarks, while FSAF leads on Levy20. These results show that even sparse, non-extendable records can provide useful experience for feedback-conditioned optimization on new tasks, helping make better use of limited, costly objective evaluations.

This setting matters when past campaigns cannot be extended: the practice used for policy learning is generated computationally from their fixed records, while target-task evaluations are reserved for deployment. The policy then uses its learned parameters together with the observations collected on the current task, without updating those parameters between queries. Our evidence concerns new tasks related to the recorded ones and sharing their decision space; it does not establish performance on unrelated task families. Within that scope, the results support learning a reusable query policy from a small historical archive rather than requiring fresh interactions with the historical objectives.

\subsection*{AI use statement}
Generative AI tools were used to assist with language polishing of the manuscript, code optimization, and organization of figures and research materials. The authors reviewed the AI-assisted work and take responsibility for the final manuscript and research results.


\newpage
\appendix
\section{Method Implementation}
\label{app:method_details}

\subsection{Practice-Function Construction}

Flatten the continuous-response historical observations and let $h_i$ identify the task associated with record $i$. The model in Eq.~\eqref{eq:joint_task_model} gives the covariance
\begin{equation}
C_{ij}=k_s(x_i,x_j)+\mathbf 1[h_i=h_j]\{k_u(x_i,x_j)+v_a\}
+\sigma_\epsilon^2\mathbf 1[i=j].
\label{eq:method_historical_covariance}
\end{equation}
Both spatial kernels are isotropic Mat\'ern-$5/2$ kernels. A new task shares only $g$ with historical tasks; its task-specific process $u_\star$ and offset $a_\star$ are sampled anew. Model parameters are fitted using the corresponding historical records, without choosing them from target-task responses or results.

We approximate prior paths with random Fourier features and condition whole paths on the historical observations. Let $z_{\mathcal D}$ be the transformed historical responses, $z_{\mathcal D}^{0}$ the prior draw at historical inputs (including the historical tasks' processes, offsets, and fitting noise), and $z_\star^{0}(x)$ the new task's prior latent path. With $c_s(x)_i=k_s(x,x_i)$, the continuous practice path, conditional on a mean and response-transform component, is
\begin{equation}
\widetilde z_\star(x)=\beta+z_\star^{0}(x)
+c_s(x)^\top C^{-1}
\left[z_{\mathcal D}-\beta\mathbf 1-z_{\mathcal D}^{0}\right].
\end{equation}
Applying the inverse response transformation, and any applicable fixed discrete response component, yields $\widetilde f(x)$. The path, transformation component, and discrete component remain fixed within a draw; historical fitting noise is not resampled on practice queries.

The joint model and path-conditioning procedure are shared across experiments. Response transformations and an optional discrete component accommodate different response distributions in the fixed historical records. Synthetic experiments use an affine transformation; engineering experiments use record-fitted monotone transformations, with a discrete component where repeated responses occur. These choices are fixed before target-task evaluation, with numerical parameters fitted only from the corresponding records; they do not alter the candidate policy, training objective, or deployment procedure.

\subsection{Online Representation, Candidates, and Training}

The online representation $\phi_t$ in Eq.~\eqref{eq:online_response_representation} serves decision inputs, whereas $T_\eta$ serves practice-function construction. Synthetic runs use $\phi_t(y_i;O_t)=y_i$. In engineering runs, let $N_t=|O_t|$ and let $r_i$ be the one-based average rank of $y_i$ among the observed responses, with ties sharing their average rank. Then
\begin{equation}
\phi_t(y_i;O_t)=\Phi^{-1}\!\left(\frac{r_i-0.5}{N_t}\right).
\end{equation}
This representation is recomputed from current observations at every step, with the same rule used in training and deployment. It leaves raw-response incumbents, rewards, and evaluation unchanged. The wide range of historical engineering responses motivates representing their relative differences without introducing hidden task parameters.

Each decision uses $M=256$ candidates: 128 points from an initial persistent scrambled Sobol pool and local points filling the remaining slots. Selected persistent points are removed. Local points are generated around up to three spatially separated, favorably observed locations. Their centers change with feedback, while the radius and allocation follow a predefined budget-dependent rule and are not policy actions. Context tokens contain locations and response-order features; candidate tokens contain locations, TabPFN predictive means and standard deviations in the online representation space, and a persistent-pool indicator. Separate attention streams encode context and candidates before candidate-to-context attention forms action scores.

The candidate construction, network design, and training objective are shared across experiments. The auxiliary-label parameters in Eq.~\eqref{eq:adaptive_labels} are $k=20$, $\alpha=0.35$, $\epsilon_\tau=10^{-6}$, and $\zeta=0.001$; the value, entropy, and auxiliary coefficients in Eq.~\eqref{eq:joint_policy_loss} are $c_V=0.5$, $c_{\mathrm{ent}}=0.01$, and $\lambda_{\mathrm{aux}}=0.2$. PPO uses complete-trajectory returns without discounting.

\section{Experimental Protocol}
\label{app:response_configurations}

This appendix expands the experimental setup in Section~\ref{sec:exp_setup}. The practice-function sampler and policy implementation are described in Appendix~\ref{app:method_details}; the benchmark task generators and evaluators are separate from that training procedure.

\subsection{Task Archives and Evaluation Budgets}

Each benchmark provides a fixed archive of $K=20$ related historical tasks, with $5d$ observed input--response pairs per task. The five synthetic families each have 100 target tasks with identities distinct from those in the corresponding historical archive. BOPTEST and SU2 each have 20 historical and 20 target tasks. Decision variables are mapped to $[0,1]^d$ within each benchmark. The dimensions are $d=6$ for Hartmann6, Ackley6, Levy6, and BOPTEST; $d=10$ for SU2; and $d=20$ for Ackley20 and Levy20.

At each target task, the methods receive the same initial design of $n_0=d/2$ scrambled Sobol points and a total budget of $B=5d$ objective evaluations, including that design. Thus $(n_0,B)$ is $(3,30)$, $(5,50)$, or $(10,100)$ for 6D, 10D, or 20D tasks, respectively. The reported target-task runs use evaluation seed 42. Every target-task objective evaluation, including the initial design, counts toward $B$; constructing and querying practice functions during training are computational operations. Once an archive is frozen, no method can request additional evaluations of a historical task. The synthetic reference values used for scoring are withheld from all optimizers and are defined in Appendix~\ref{app:synthetic_results}.

\subsection{Synthetic Task Construction}
\label{app:synthetic_construction}

Hartmann6, Ackley6, Levy6, Ackley20, and Levy20 denote related-task families, not repeated optimization of a single unmodified benchmark function. All tasks are defined on $[0,1]^d$. A benchmark generator samples each task's parameters and provides objective responses to the historical logger or the target-task evaluator; neither the parameters nor unqueried responses are available to CAP training. Historical and target tasks have distinct identities, and all 100 sampled targets in each family are retained without selection by method performance.

\paragraph{Task variation.}
Hartmann6 retains the four basin centers $P_i$ and base precision coefficients $A_{ij}$ of the Hartmann function. For task parameters $\theta$, its objective has the form
\begin{equation}
f_\theta(x)=s_\theta\left[-\sum_{i=1}^{4}\alpha_i^{(\theta)}
\exp\!\left(-\sum_{j=1}^{6}A_{ij}m_i^{(\theta)}c_j^{(\theta)}
\bigl(W_\theta(x)_j-P_{ij}\bigr)^2\right)\right].
\end{equation}
Here $W_\theta$ is a boundary-preserving invertible warp, while $\alpha_i^{(\theta)}$, $m_i^{(\theta)}$, and $c_j^{(\theta)}$ vary basin depth, width, and coordinate anisotropy. The base depths are $(1,1.2,3,3.2)$. Warp shifts, scales, and pairwise coupling angles range over $[-0.0354,0.0354]$, $[0.9056,1.0944]$, and $[-11.8^\circ,11.8^\circ]$, respectively; depth factors range over $[0.5,1.5]$. Width and anisotropy receive smaller task-specific perturbations. The generator-only factor $s_\theta$ keeps the numerical reference minimum on the base Hartmann scale. Neither $s_\theta$ nor the task parameters are supplied to an optimizer.

For Ackley and Levy, a task combines a base function, the same type of invertible warp, and a localized smooth perturbation. Let $R$ map unit-cube coordinates to the corresponding base-function domain, coordinatewise. Then
\begin{equation}
f_\theta(x)=f_{\mathrm{base}}(R(z))+A_\theta
\exp\!\left(-\frac{\lVert z-z^{\mathrm{opt}}\rVert^2}{2L^2}\right)
\phi_\theta(z),\qquad z=W_\theta(x).
\end{equation}
Specifically, $R(z)=10z-5$ for Ackley and $R(z)=20z-10$ for Levy, giving base-function domains $[-5,5]^d$ and $[-10,10]^d$, respectively; $z^{\mathrm{opt}}$ is the corresponding base optimum in unit-cube coordinates. The localized perturbation $\phi_\theta$ uses 32 cosine features with task-specific coefficients and envelope width $L=0.18\sqrt{d/6}$. Its amplitude is $0.3$ times the family reference scale for Ackley and $0.1$ for Levy. Warp shifts, scales, and coupling angles range over $[-0.018,0.018]$, $[0.952,1.048]$, and $[-6^\circ,6^\circ]$, respectively. These variations produce distinct related tasks within a common decision space.

\subsection{Engineering Simulation Tasks}

BOPTEST evaluates a six-parameter building-control problem using the \path{multizone_office_simple_air} FMU in BOPTEST v0.9.0. Tasks correspond to different operating windows in the same building and Chicago representative weather year. Each window has a seven-day warm-up followed by three scored days at one-hour control intervals. The six decision variables are occupied heating and cooling setpoints ($18$--$22^\circ$C and $22$--$26^\circ$C), unoccupied heating and cooling setpoints ($12$--$18^\circ$C and $26$--$32^\circ$C), supply-air temperature ($12$--$18^\circ$C), and duct static pressure ($100$--$350$ Pa). A shared coordinate mapping enforces heating--cooling deadbands of at least $2^\circ$C when occupied and $4^\circ$C when unoccupied. The minimized scalar objective is $\mathrm{cost}_{\mathrm{tot}}+\mathrm{tdis}_{\mathrm{tot}}+\mathrm{idis}_{\mathrm{tot}}/494$; it is not normalized by a task-specific optimum.

SU2 evaluates ten-dimensional RAE2822 airfoil designs with SU2 v8.5.0, an SA turbulence model, fixed target lift coefficient, and cold-start RANS evaluations from a shared mesh. The design coordinates control five Hicks--Henne deformations on each of the upper and lower surfaces, each within $\pm0.001$ chord. Tasks vary flight conditions: Mach number in $[0.79,0.80]$, Reynolds number in $[6\times10^6,7\times10^6]$, and target lift coefficient in $[0.70,0.74]$. The benchmark minimizes a prespecified scalar objective: feasible simulations use $C_D/(1+C_D)$ when $C_M<0.093$ and maximum thickness $t_{\max}>0.12$; constraint violations receive $1+v/(1+v)$, where $v=\max\{0,C_M/0.093-1,1-t_{\max}/0.12\}$, and classified geometry or numerical failures receive $2$. The original scalar objectives, rather than regret to an unknown optimum, are reported for both simulators.

\subsection{Baseline Information and Training}

The single-task methods EI, PI, UCB, RolloutEI-2, FunBO, and TuRBO use the current target's observations and their native surrogates and search procedures. TAF-ranking, RGPE, ABLR, and FSBO use the fixed historical records through their transfer mechanisms. FSAF, MetaBO, and NAP are trained on the same record-derived, queryable practice-function distribution as \ours, while retaining their own predictors, candidates, policies, and learning objectives; these practice queries do not access historical or target simulators. PFNs4BO uses its released externally pretrained model. \ours{} uses a separate frozen, externally pretrained TabPFN predictor for online features and learns its candidate policy from the historical archive. All methods retain their native target-task decision procedures, and only actual target-objective evaluations are budget-matched.

\section{Synthetic Evaluation and Complete Trajectories}
\label{app:synthetic_results}

\begin{figure}[!htbp]
\centering
\includegraphics[page=1,width=0.90\linewidth]{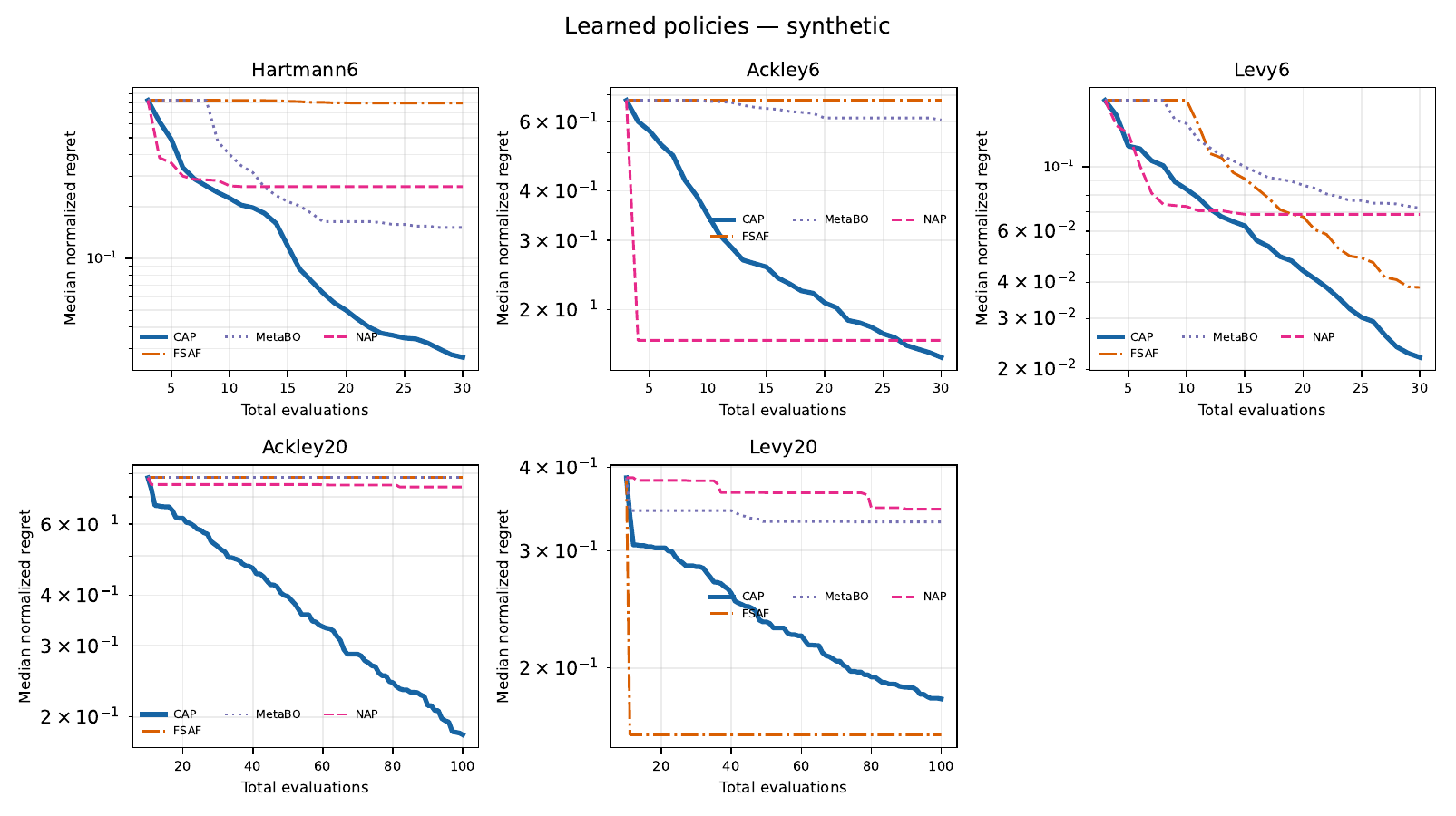}
\caption{Median normalized simple regret for \ours{} and the learned query policies FSAF, MetaBO, and NAP across the five synthetic families. Lower is better; vertical scales differ by family.}
\label{fig:synthetic_learned_curves}
\end{figure}

For target task $i$, let $b_{i,t}$ be the lowest objective value observed through evaluation $t$. Given the task's reference minimum $m_i^{\mathrm{ref}}$ and maximum $M_i^{\mathrm{ref}}$ for evaluation, normalized simple regret is
\[
  r_{i,t}=\frac{b_{i,t}-m_i^{\mathrm{ref}}}
  {M_i^{\mathrm{ref}}-m_i^{\mathrm{ref}}}.
\]
The reference values are used only for offline scoring and are not supplied to the optimizers; reference minima obtained by finite search are not claimed to be certified global optima. At each evaluation count, we take the median of $r_{i,t}$ across the 100 target tasks in a family. The horizontal axis includes evaluations used by the initial design, so trajectories begin once that design is complete. Table~\ref{tab:synthetic_median_regret} reports final-budget values; Figures~\ref{fig:synthetic_learned_curves}--\ref{fig:synthetic_single_ii_curves} give the complete trajectories by method category under the protocol in Appendix~\ref{app:response_configurations}.

\begin{figure}[!htbp]
\centering
\includegraphics[page=3,width=0.90\linewidth]{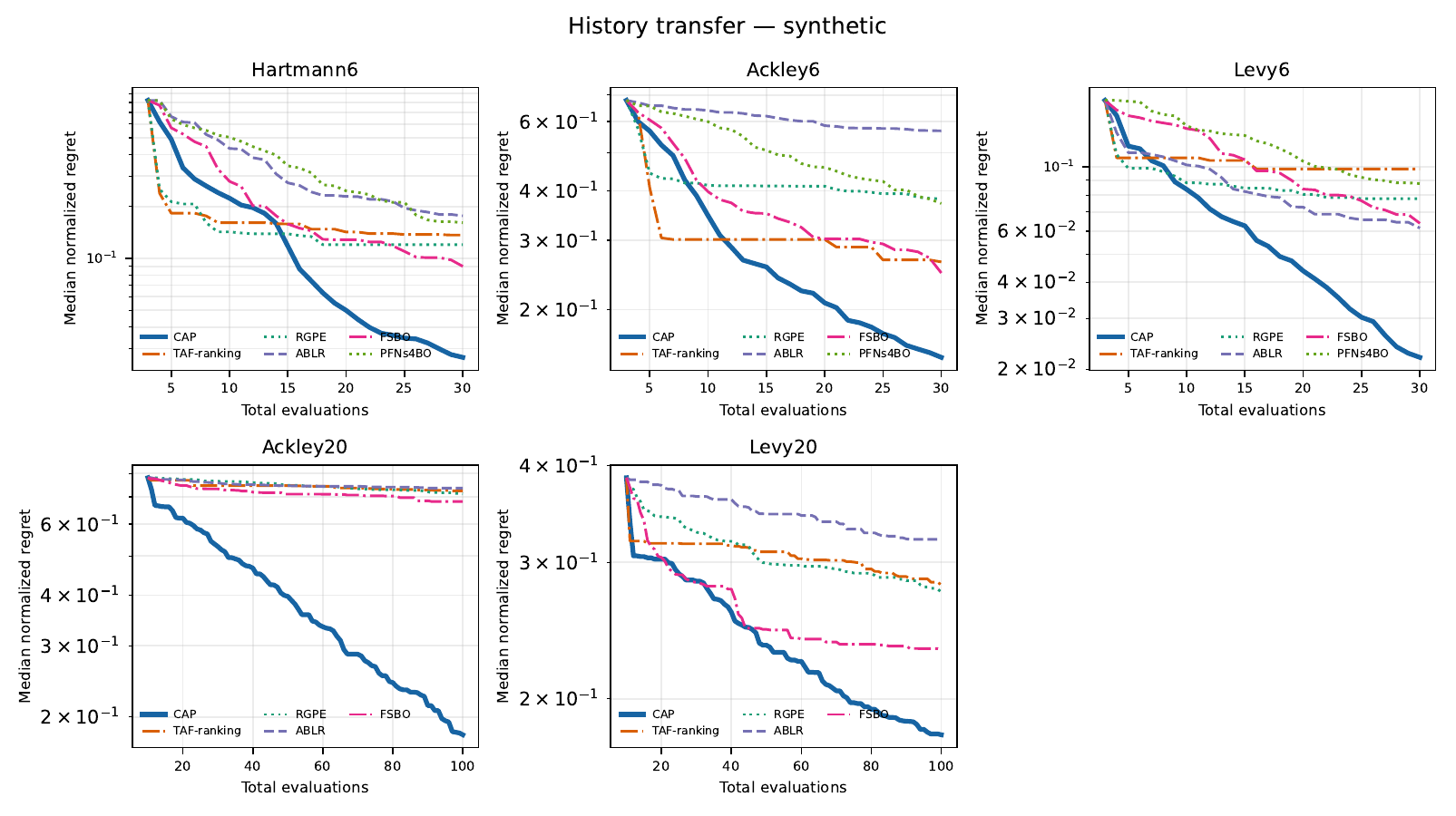}
\caption{Median normalized simple regret for \ours{}, TAF-ranking, RGPE, ABLR, FSBO, and PFNs4BO. The released PFNs4BO model supports at most 18 input dimensions, so it has no 20D curves.}
\label{fig:synthetic_transfer_curves}
\end{figure}

\begin{figure}[!htbp]
\centering
\includegraphics[page=5,width=0.90\linewidth]{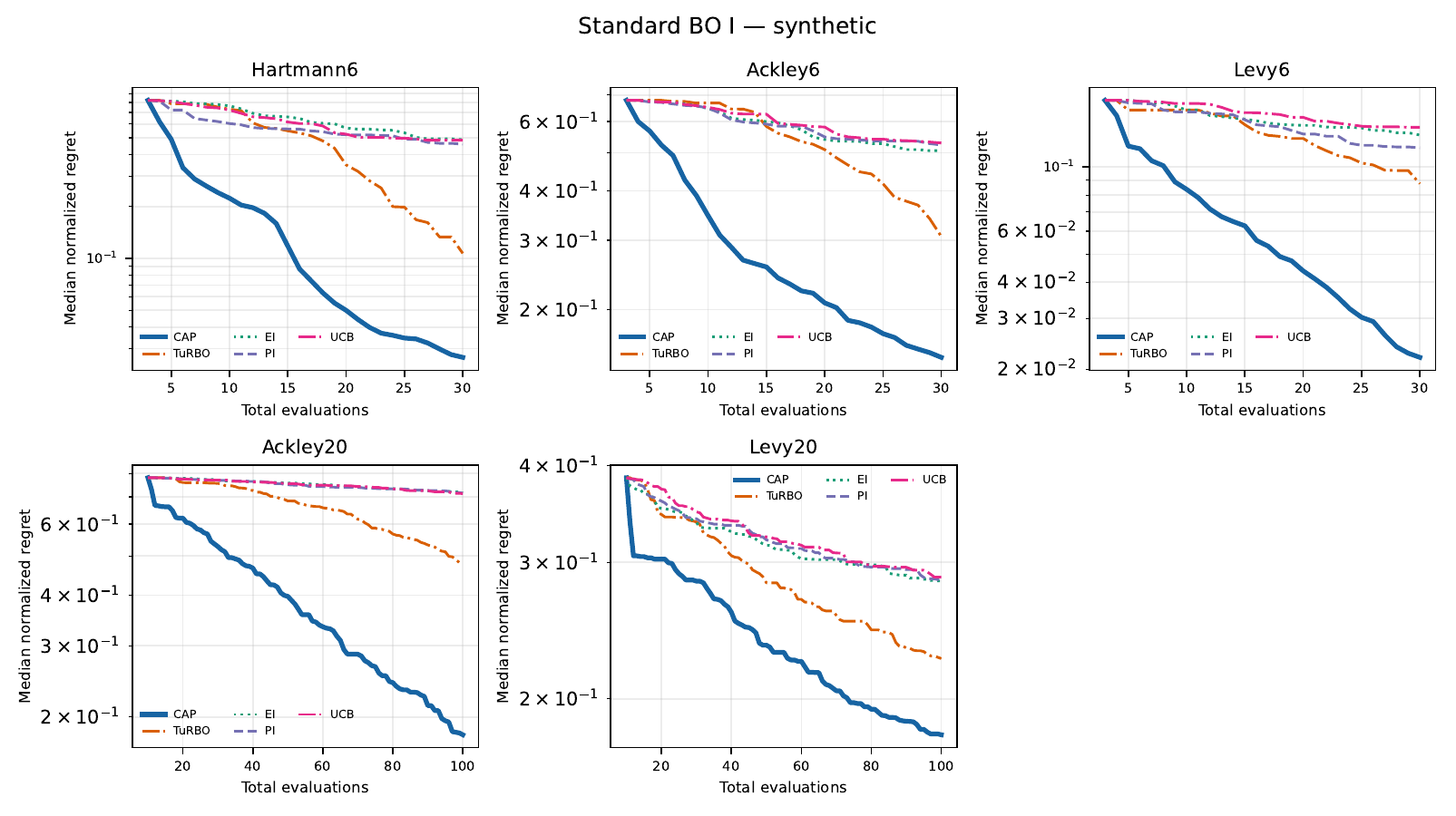}
\caption{Median normalized simple regret for \ours{}, TuRBO, EI, PI, and UCB. The evaluation count includes the initial design.}
\label{fig:synthetic_single_i_curves}
\end{figure}

\begin{figure}[!htbp]
\centering
\includegraphics[page=7,width=0.90\linewidth]{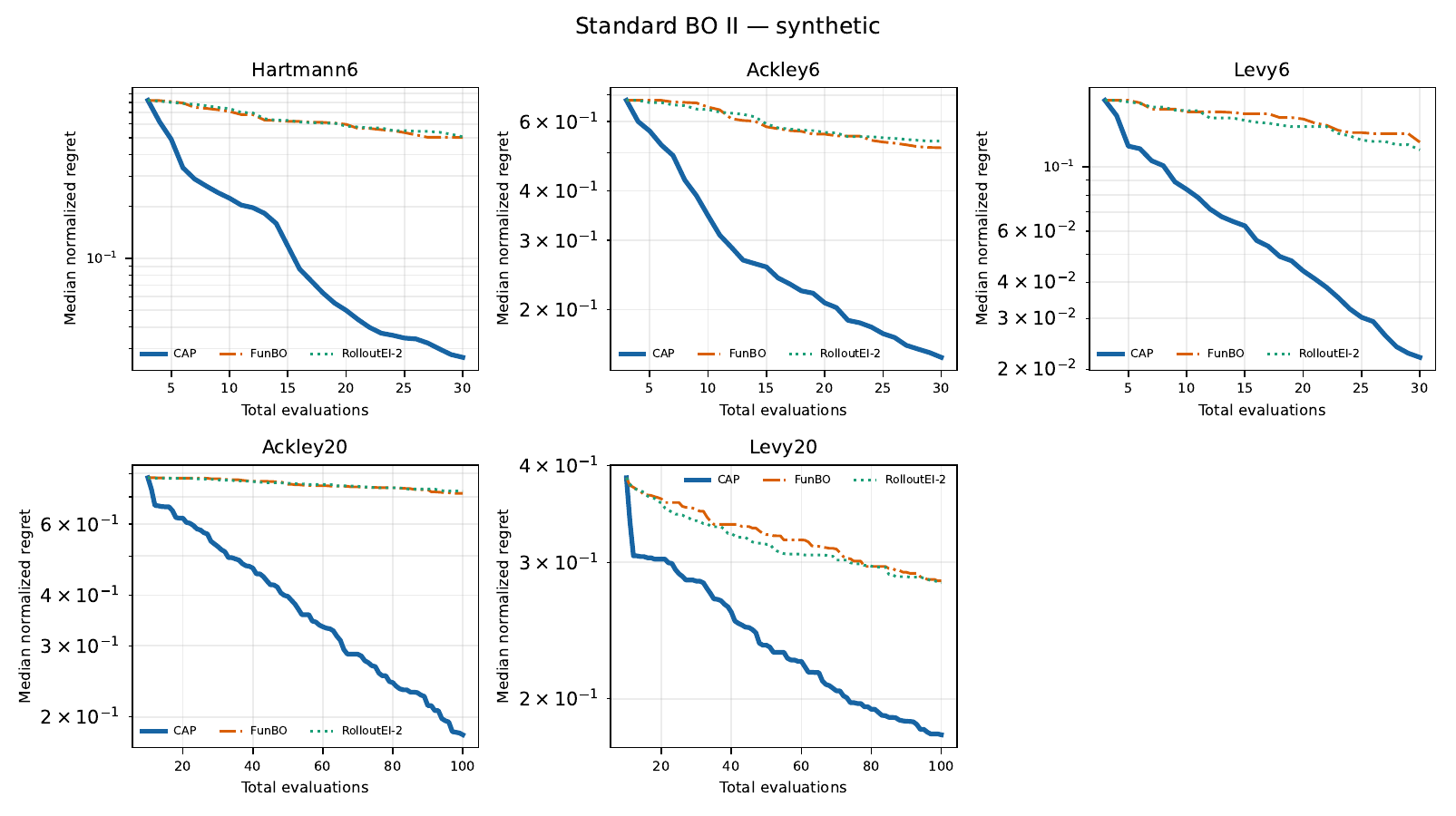}
\caption{Median normalized simple regret for \ours{}, FunBO, and RolloutEI-2. Each point is a cross-task median at the indicated evaluation count.}
\label{fig:synthetic_single_ii_curves}
\end{figure}

\FloatBarrier

\section{Complete Engineering Simulation Trajectories}
\label{app:engineering_curves}

\begin{figure}[!htbp]
\centering
\includegraphics[page=2,width=0.90\linewidth]{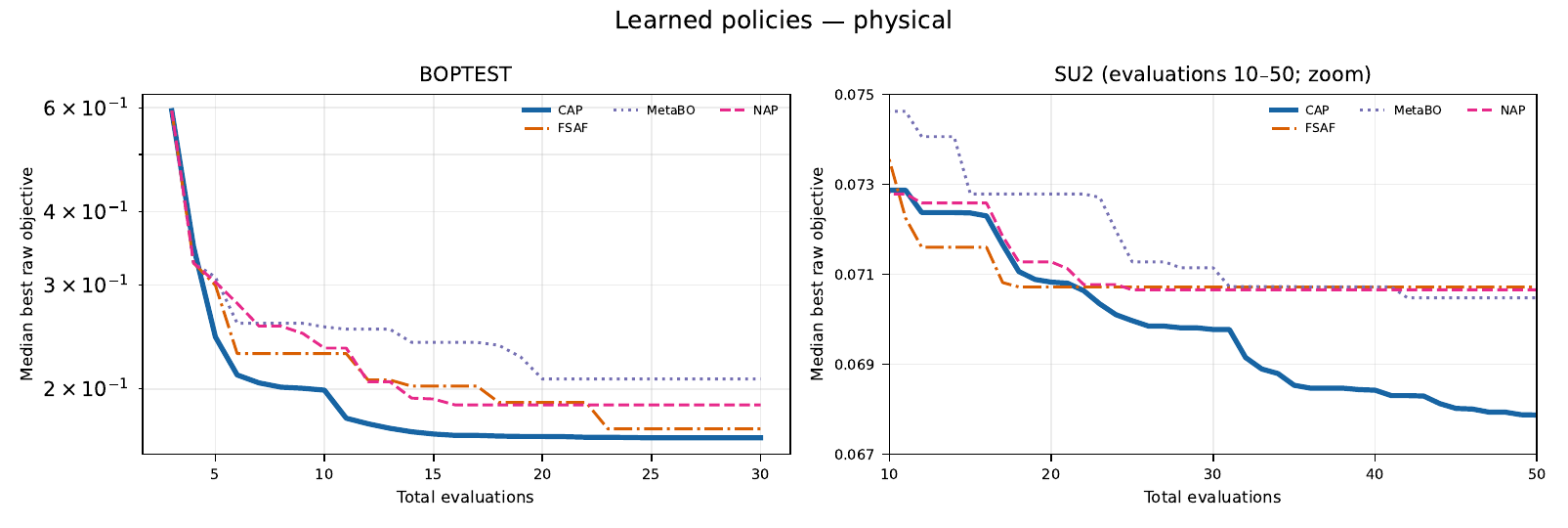}
\caption{Median best-observed objective for \ours{} and the learned query policies FSAF, MetaBO, and NAP on BOPTEST and SU2.}
\label{fig:engineering_learned_curves}
\end{figure}

\begin{figure}[!htbp]
\centering
\includegraphics[page=4,width=0.90\linewidth]{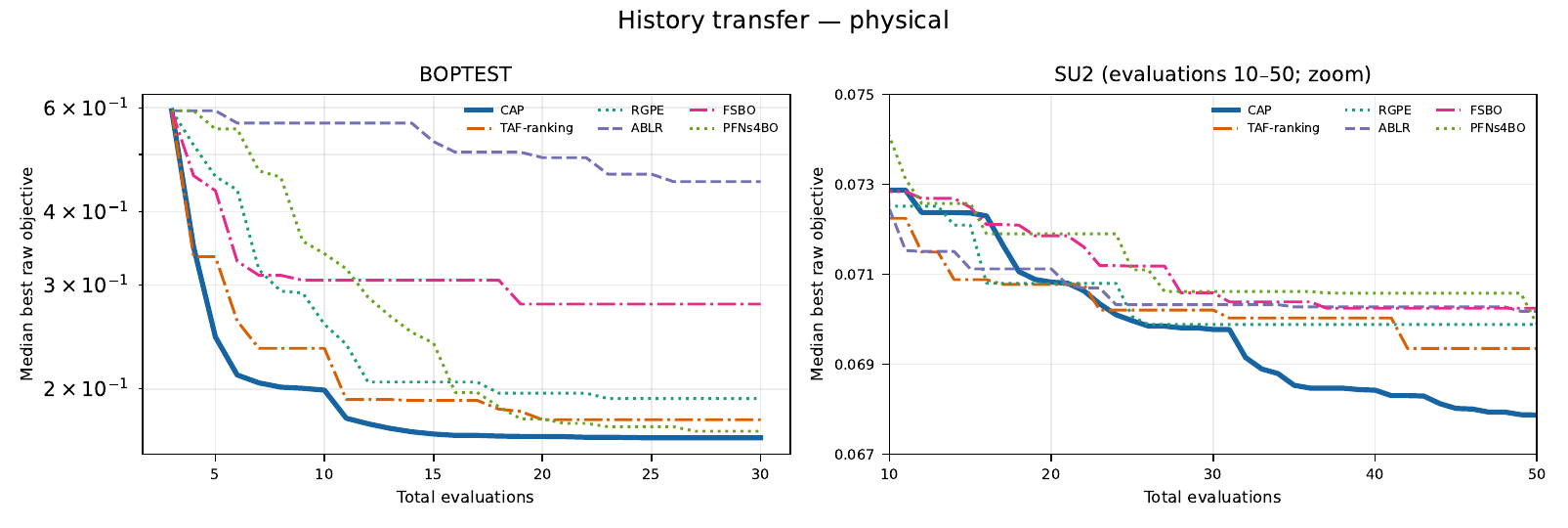}
\caption{Median best-observed objective for \ours{}, TAF-ranking, RGPE, ABLR, FSBO, and PFNs4BO on both engineering benchmarks.}
\label{fig:engineering_transfer_curves}
\end{figure}

\begin{figure}[!htbp]
\centering
\includegraphics[page=6,width=0.90\linewidth]{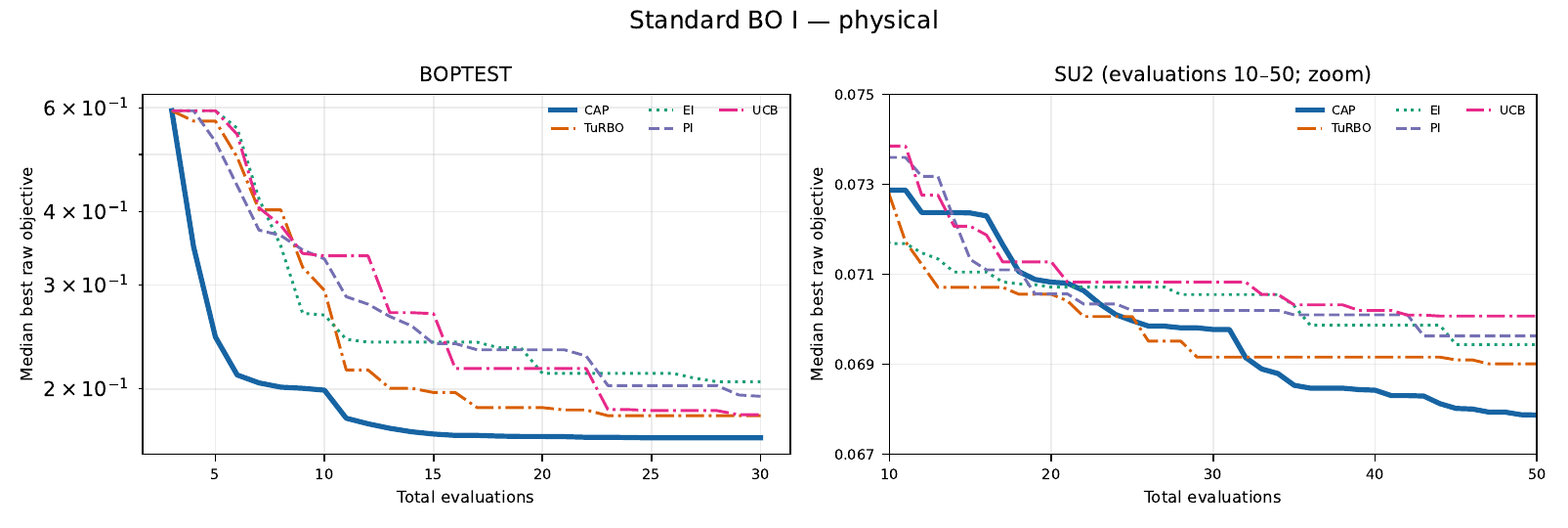}
\caption{Median best-observed objective for \ours{}, TuRBO, EI, PI, and UCB on both engineering benchmarks.}
\label{fig:engineering_single_i_curves}
\end{figure}

\begin{figure}[!htbp]
\centering
\includegraphics[page=8,width=0.90\linewidth]{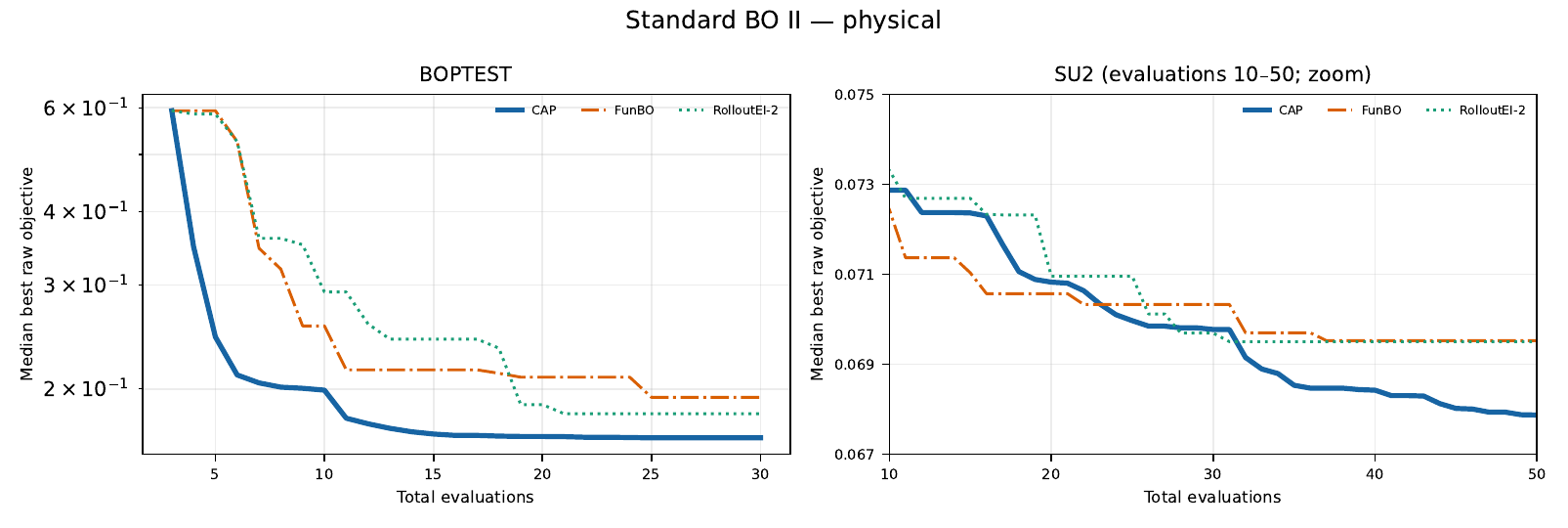}
\caption{Median best-observed objective for \ours{}, FunBO, and RolloutEI-2 on both engineering benchmarks.}
\label{fig:engineering_single_ii_curves}
\end{figure}

The engineering benchmarks have no known global optimum from which to compute regret, so we use the original objective, as in the main text. For target task $i$, let $y_{i,s}$ be the objective value at simulator evaluation $s$ and $b_{i,t}=\min_{1\leq s\leq t}y_{i,s}$ the best value observed through evaluation $t$. At each $t$, the curves report the median of $b_{i,t}$ over the 20 target tasks; lower is better. Evaluation counts include the initial designs of three points for BOPTEST and five for SU2. The SU2 panels display evaluations 10--50 on a linear, truncated vertical axis spanning 0.067--0.075 to resolve the late-budget differences; earlier evaluations still count toward the budget. Table~\ref{tab:physical_final} gives the final-budget comparison, while Figures~\ref{fig:engineering_learned_curves}--\ref{fig:engineering_single_ii_curves} show every reported method by category.

\end{document}